\documentclass[11pt,a4paper]{article}
\usepackage[hyperref]{acl2021}
\usepackage{times}
\usepackage{latexsym}

\usepackage{microtype}

\aclfinalcopy 
\def\aclpaperid{3154} 

\usepackage{times}
\usepackage{latexsym}

\usepackage{stmaryrd}
\usepackage{graphicx}
\usepackage{amsmath}
\usepackage{xcolor}
\usepackage{soul}
\usepackage{amsfonts}
\usepackage{graphicx}
\usepackage{hyperref}
\usepackage{caption}
\usepackage{subcaption}
\usepackage{multirow}
\usepackage{adjustbox}

\DeclareRobustCommand{\hlcolor}[1]{{\sethlcolor{lightgray}\hl{#1}}}

\usepackage[T1]{fontenc}

\usepackage[utf8]{inputenc}

\title{Multi-Task Learning and Adapted Knowledge Models for Emotion-Cause Extraction}

\author{Elsbeth Turcan$^1$\thanks{~~Work done during an internship with Amazon AI.} \\ \textbf{Kasturi Bhattacharjee}$^3$
	\\\And Shuai Wang$^3$ \\ \textbf{Yaser Al-Onaizan}$^3$ \\ $^1$Department of Computer Science, Columbia University \\ $^2$Data Science Institute, Columbia University \\ $^3$Amazon AI \\ \texttt{\{eturcan, smara\}@cs.columbia.edu} \\ \texttt{\{wshui, ranubhai, kastb, onaizan\}@amazon.com} \\\And Rishita Anubhai$^3$ \\ \textbf{Smaranda Muresan}$^{2,3}$}

\begin{document}
\maketitle
\begin{abstract}
Detecting what emotions are expressed in text is a well-studied problem in natural language processing. However, research on finer-grained emotion analysis such as what causes an emotion is still in its infancy. We present solutions that tackle both emotion recognition and emotion cause detection in a joint fashion. Considering that common-sense knowledge plays an important role in understanding implicitly expressed emotions and the reasons for those emotions, we propose novel methods that combine common-sense knowledge via adapted knowledge models with multi-task learning to perform joint emotion classification and emotion cause tagging. We show performance improvement on both tasks when including common-sense reasoning and a multi-task framework. We provide a thorough analysis to gain insights into model performance.
\end{abstract}

\section{Introduction}

Utterance and document level emotion recognition has received significant attention from the research community \cite{SemEval2018Task1,poria2020beneath}.  Given the utterance \emph{Sudan protests: Outrage as troops open fire on protestors} an emotion recognition system will be able to detect that \emph{anger} is the main expressed emotion, signaled by the word "outrage".  However, the semantic information associated with expressions of emotion, such as the cause (the thing that triggers the emotion) or the target (the thing toward which the emotion is directed), is important to provide a finer-grained understanding of the text that might be needed in real-world applications. In the above utterance, the cause of the anger emotion is the event ``troops open fire on protestors'', while the target is the entity "troops" (see \autoref{fig:goodnewseveryone}). 

Research on finer-grained emotion analysis such as detecting the cause for an emotion expressed in text is in its infancy. Most work on emotion-cause detection has utilized a Chinese dataset where the cause is always syntactically realized as a clause and thus was modeled as a classification task \citep{gui-etal-2016-event}. However, recently \citet{DBLP:conf/lrec/BostanKK20} and \citet{oberlander-klinger-2020-token} argued that in English, an emotion cause can be expressed syntactically as a clause (\textit{as troops open fire on protestors}), noun phrase (\textit{1,000 non-perishable food donations}) or verb phrase (\textit{jumped into an ice-cold river}), and thus we follow their approach of framing emotion cause detection as a sequence tagging task. 

We propose several ways in which to approach the tasks of emotion recognition and emotion cause tagging. First, these two tasks should not be independent; because the cause is the trigger for the emotion, knowledge about what the cause is should narrow down what emotion may be expressed, and vice versa. Therefore, we present a multi-task learning framework to model them jointly. Second, considering that common-sense knowledge plays an important role in understanding implicitly expressed emotions and the reasons for those emotions, we explore the use of common-sense knowledge via adapted knowledge models  (COMET, \citet{DBLP:conf/acl/BosselutRSMCC19}) for both tasks. A key feature of our approach is to combine these adapted knowledge models (i.e., COMET), which are specifically trained to use and express common-sense knowledge, with pre-trained language models such as BERT, \citep{DBLP:conf/naacl/DevlinCLT19}. 

Our primary contributions are three-fold: (i) an under-studied formulation of the emotion cause detection problem as a sequence tagging problem; (ii) a set of models that perform the emotion classification and emotion cause tagging tasks jointly while using common-sense knowledge (\autoref{sec:multitask}) with improved performance (\autoref{sec:results}); and (iii) analysis to gain insight into both model performance and the GoodNewsEveryone dataset that we use \citep{DBLP:conf/lrec/BostanKK20} (\autoref{sec:analysis}). 

\section{Related Work}

Emotion detection is a widely studied subfield of natural language processing \citep{SemEval2018Task1,poria2020beneath}, and has been applied to a variety of text genres such as fictional stories \citep{10.3115/1220575.1220648}, news headlines \citep{DBLP:series/sci/StrapparavaM10}, and social media, especially microblogs such as Twitter \citep{abdul-mageed-ungar-2017-emonet,10.5555/2693068.2693087,DBLP:journals/corr/abs-1912-02387,SemEval2018Task1}. Earlier work, including some of the above, focused on feature-based machine learning models that could leverage emotion lexicons \citep{DBLP:journals/corr/MohammadT13}), while recent work explores deep learning models (e.g., Bi-LSTM and BERT) and multi-task learning \citep{DBLP:journals/corr/abs-1809-04505,demszky-etal-2020-goemotions}.

However, comparatively few researchers have looked at the semantic roles related to emotion such as the cause, the target or the experiencer, with few exceptions for Chinese \cite{gui-etal-2016-event,chen-etal-2018-joint,DBLP:journals/corr/abs-1906-01267,DBLP:journals/corr/abs-1906-01236,fan-etal-2020-transition,wei-etal-2020-effective,ding-etal-2020-ecpe}, English \cite{mohammad-etal-2014-semantic,Ghazi2015DetectingES,kim-klinger-2018-feels,DBLP:conf/lrec/BostanKK20, oberlander-etal-2020-experiencers,oberlander-klinger-2020-token} and Italian \cite{russo-etal-2011-emocause}. 
We highlight some of these works here and draw connection to our work. Most recent work on emotion-cause detection has been carried out on a Chinese dataset compiled by \citet{gui-etal-2016-event}. This dataset characterizes the emotion and cause detection problems as clause-level pair extraction problem -- i.e., of all the clauses in the input, one is selected to contain the expression of an emotion, and one or more (usually one) are selected to contain the cause of that emotion. Many publications have used this corpus to develop novel and effective model architectures for the clause-level classification problem \citep{chen-etal-2018-joint,DBLP:journals/corr/abs-1906-01267,DBLP:journals/corr/abs-1906-01236,fan-etal-2020-transition,wei-etal-2020-effective,ding-etal-2020-ecpe}. The key difference between this work and ours is that we perform cause detection as a sequence-tagging problem: the cause may appear anywhere in the input, and may be expressed as any grammatical construction (a noun phrase, a verb phrase, or a clause). Moreover, we use common sense knowledge for both tasks (emotion and cause tagging), through the use of adapted language models such as COMET.

For English, several datasets have been introduced \cite{mohammad-etal-2014-semantic, kim-klinger-2018-feels,Ghazi2015DetectingES,DBLP:conf/lrec/BostanKK20,poria2020recognizing}, and emotion cause detection has been tackled either as a classification problem \cite{mohammad-etal-2014-semantic}, or as a sequence tagging or span detection problem \cite{kim-klinger-2018-feels,Ghazi2015DetectingES,oberlander-klinger-2020-token,poria2020recognizing}. We particularly note the work of \citet{oberlander-klinger-2020-token}, who argue for our problem formulation of cause detection as sequence tagging rather than as a classification task supported by empirical evidence on several datasets including the GoodNewsEveryone dataset \cite{DBLP:conf/lrec/BostanKK20} we use in this paper. One contribution we bring compared to these models is that we formulate a multi-task learning framework to jointly learn the emotion and the cause span. Another contribution is the use of common-sense knowledge through the use of adapted knowledge models such as COMET (both in the single models and the multi-task models). \newcite{ghosal-etal-2020-cosmic} have very recently shown the usefulness of common-sense reasoning to the task of conversational emotion detection. 

\section{Data}

\begin{figure}[t]
	\begin{center}
		\includegraphics[scale=0.5]{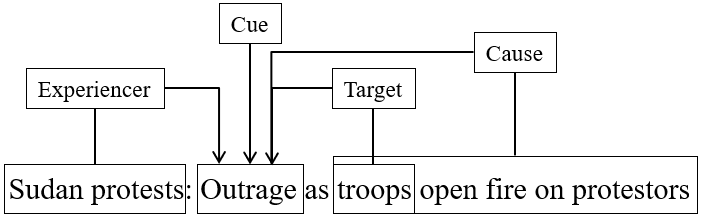}
	\end{center}
	\caption{An example of the semantic roles annotated by \citet{DBLP:conf/lrec/BostanKK20}}
	\label{fig:goodnewseveryone}
\end{figure}

\begin{figure}[t]
	\includegraphics[scale=0.5]{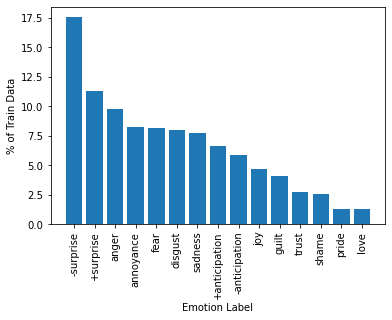}
	\caption{Distribution of adjudicated emotion labels in the GoodNewsEveryone train data, as a percentage of the data points. ``Positive'' and ``Negative'' are abbreviated as + and -.}
	\label{fig:label-distribution}
\end{figure}

\begin{figure*}
    \begin{subfigure}[t]{0.5\textwidth}
        \centering
        \includegraphics[scale=0.35]{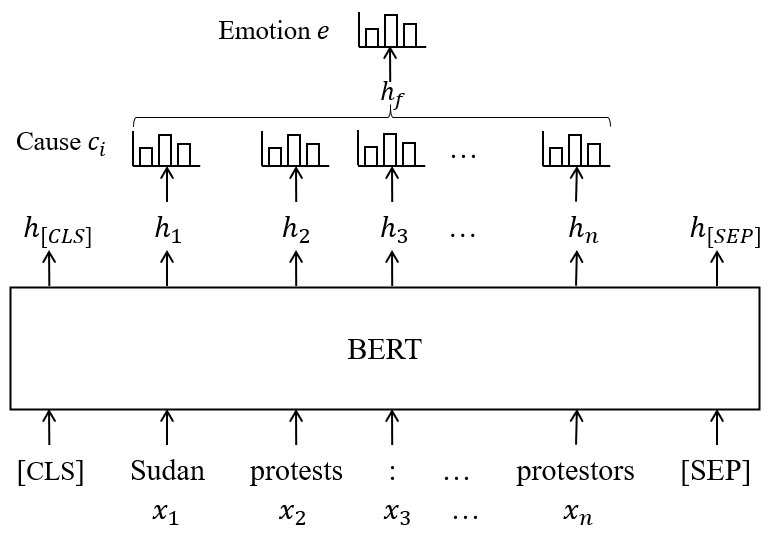} 
        \caption{The $\text{Multi}_{C \shortrightarrow E}$ model.} \label{fig:causefirst-arch}
    \end{subfigure}
    \begin{subfigure}[t]{0.5\textwidth}
        \centering
         \includegraphics[scale=0.36]{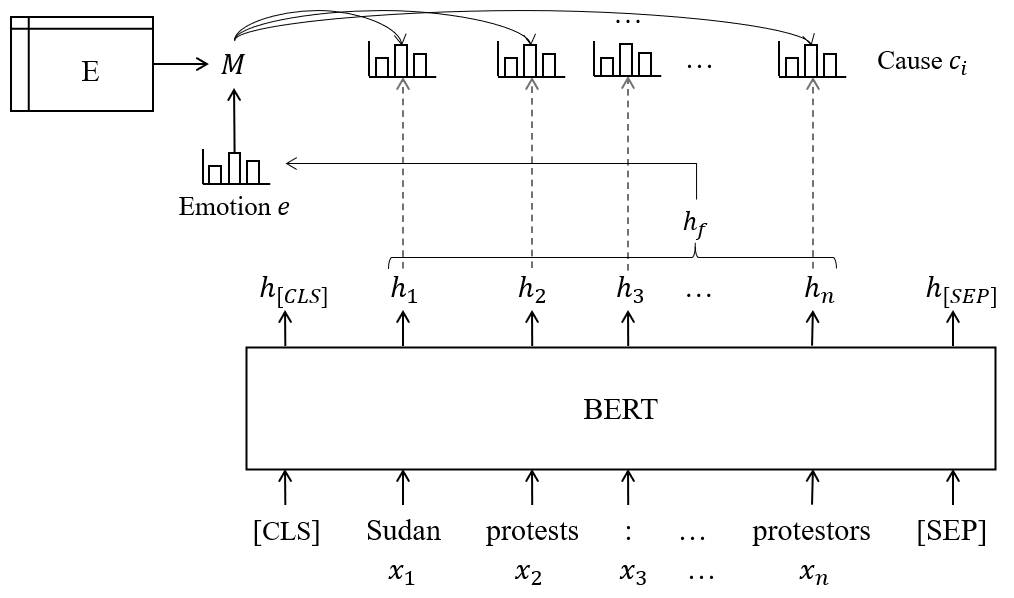}
        \caption{The $\text{Multi}_{E \shortrightarrow C}$ model.} \label{fig:emofirst-arch}
    \end{subfigure}
    \caption{Our multi-task models.}
\end{figure*}

For our experiments, we use the GoodNewsEveryone corpus \citep{DBLP:conf/lrec/BostanKK20}, which contains 5,000 news headlines labeled with emotions and semantic roles such as the target, experiencer, and cause of the emotion, as shown in \autoref{fig:goodnewseveryone}.\footnote{While the dataset labels both the most dominant emotion expressed in text and the reader's emotion, for this paper we only focus on the former.}  
We focus on the emotion detection and cause tagging tasks in this work. To our knowledge, GoodNewsEveryone is the largest English dataset labeled for both of these tasks.

In our experiments, we limit ourselves to the data points for which a cause span was annotated (4,798). We also note that this dataset uses a 15-way emotion classification scheme, an extended set including the eight basic Plutchik emotions as well as additional emotions like \textit{shame} and \textit{optimism}. While a more fine-grained label set is useful for capturing subtle nuances of emotion, many external resources focus on a smaller set of emotions. We also note that the label distribution of this dataset heavily favors the more basic emotions, as shown in \autoref{fig:label-distribution}. Therefore, for our work, we choose to limit ourselves to the six Ekman emotions (\textit{anger}, \textit{fear}, \textit{disgust}, \textit{joy}, \textit{surprise}, and \textit{sadness}). We also choose to keep \textit{positive surprise} and \textit{negative surprise} separated, to avoid severely unbalancing the label distribution for our experiments. We randomly split the remaining data (2,503 data points) into 80\% train, 10\% development, and 10\% test.

\section{Models} \label{sec:models}

An important feature showcased by the GoodNewsEveryone dataset is that causes of emotions can be expressed through different syntactic constituents such as clauses, verb phrases, or noun-phrases. Thus, we approach the cause detection problem as a sequence tagging problem using the IOB scheme \citep{DBLP:journals/corr/cmp-lg-9505040}:
$\mathcal{C} = \{\text{I-cause}, \text{O}, \text{B-cause}\}$. Our approach is supported by very recent results by \citet{oberlander-klinger-2020-token} and \citet{yuan-etal-2020-emotion} who show that modeling emotion cause detection as a sequence tagging problem is better suited than a clause classification problem, although not much current work has yet adopted this formulation.  We tackle the emotion detection task as a seven-way classification task with $\mathcal{E} = \{\text{anger}, \text{disgust}, \text{fear}, \text{joy}, \text{sadness}, \text{negative}$ $\text{surprise}, \text{positive surprise}\}$. 

\subsection{Single-Task Models} \label{sec:singletask}

As a baseline, we train single-task models for each of emotion classification and cause span tagging. We use a pre-trained BERT language model\footnote{We use \textsc{BERT-base-uncased}. We experimented with \textsc{BERT-base-cased}, but it underperformed as the headlines incorporated into GoodNewsEveryone come from different news sources and have different capitalization styles.}  \citep{DBLP:conf/naacl/DevlinCLT19}, which we fine-tune on our data, as the basis of this model. Our preprocessing strategy for all of our models consists of the pretrained BERT vocabulary and WordPiece tokenizer\footnote{In the tagging setting, we ignore all tags predicted for subword tokens and use only the tag of the first subword.} \citep{DBLP:journals/corr/WuSCLNMKCGMKSJL16} from Huggingface \citep{wolf-etal-2020-transformers}. Therefore, for a sequence of $n$ WordPiece tokens, our input to the BERT model is a sequence of $n + 2$ tokens, $X = [[\text{CLS}], x_1, x_2, ... x_n, [\text{SEP}]]$, where each $x_i$ is from a finite WordPiece vocabulary and [CLS] and [SEP] are BERT's begin and end tokens. Passing $X$ through BERT yields a sequence of vector hidden states $H = [h_{[CLS]}, h_1, h_2, ..., h_n, h_{[SEP]}]$ with dimension $d_{BERT} = 768$. For emotion classification, we pool these hidden states and allow hyperparameter tuning to select the best type: selecting the [CLS] token ($h_f = h_{[CLS]}$), mean pooling ($h_f = \frac{\sum_{i=1}^n h_i}{n}$), max pooling ($h_{f,j} = \max{h_{i,j}}$), or attention as formulated by \citet{DBLP:journals/corr/BahdanauCB14}:

\begin{equation} \label{eqn:bahdanau-attn}
	h_f = \sum^{n}_{i=1} \alpha_i h_i
\end{equation}

where $\alpha_i = \frac{\exp{(W_ah_i + b_a)}}{\sum_{j=1}^n \exp{(W_ah_j + b_a)}}$ for trainable weights $W_a \in \mathbb{R}^{1 \times d_{BERT}}$ and $b_a \in \mathbb{R}^{1}$. Then, the final distribution of emotion scores is calculated by a single dense layer and a softmax: 

\begin{equation} \label{eqn:emotion}
	e = \text{softmax}(W_eh_f + b_e)
\end{equation}

with $e \in \mathbb{R}^{|\mathcal{E}|}$ and for trainable parameters $W_e \in \mathbb{R}^{|\mathcal{E}| \times d_{BERT}}$ and $b_e \in \mathbb{R}^{|\mathcal{E}|}$. For cause tagging, a tag probability distribution is calculated directly on each hidden state: 

\begin{equation} \label{eqn:cause}
	c_i = \text{softmax} (W_ch_i + b_c)
\end{equation}

with $c_i \in \mathbb{R}^{|\mathcal{C}|}$ and for trainable parameters $W_c \in \mathbb{R}^{|\mathcal{C}| \times d_{BERT}}$ and $b_c \in \mathbb{R}^{|\mathcal{C}|}$. We refer to both of these single-task models as BERT; if the task is not clear from the context, we will refer to the emotion detection model as $\text{BERT}_E$ and the cause tagging model as $\text{BERT}_C$. Our training loss for emotion classification as well as emotion cause tagging is the mean negative log-likelihood (NLL) loss per minibatch of size \textit{b}:

\begin{equation} \label{eqn:emo-loss}
	\text{NLL}_{\text{emo}} = - \frac{1}{b} \sum_j \sum_k y_{jk} \log e_{jk}
\end{equation}

\begin{equation} \label{eqn:cause-loss}
	\text{NLL}_{\text{cause}} = - \frac{1}{b} \sum_i \sum_j \sum_k y_{ijk} \log c_{ijk}
\end{equation}

where $j$ is the index of the sentence in the minibatch, $k$ is the index of the label being considered (emotion labels for $\text{NLL}_{\text{emo}}$ and IOB tags for $\text{NLL}_{\text{cause}}$), $i$ is the index of the $i^{th}$ token in the $j^{th}$ sentence in the minibatch, $y_{jk} \in \{0, 1\}$ is the gold probability of the $k^{th}$ emotion label for the $j^{th}$ sentence, $y_{ijk} \in \{0, 1\}$ is the gold probability of the $k^{th}$ cause tag for the $i^{th}$ token in the $j^{th}$ sentence, and $e_{jk}$ and $c_{ijk}$ are the output probabilities of the ${k^{th}}$ emotion label and of the $k^{th}$ cause label for the $i^{th}$ token, both for the $j^{th}$ sentence.

\subsection{Multi-Task Models} \label{sec:multitask}

Our hypothesis is that the emotion detection and cause tagging tasks are closely related and can inform each other; therefore we propose three multi-task learning models to test this hypothesis. For all multi-task models, we use the same base architecture (BERT) as the single models. Additionally, for these models, we combine the losses of both tasks and weight them with a tunable lambda parameter: $\lambda \text{NLL}_\text{emo} + (1 - \lambda) \text{NLL}_\text{cause}$, using $\text{NLL}_\text{emo}$ and $\text{NLL}_\text{cause}$ from \autoref{eqn:emo-loss} and \autoref{eqn:cause-loss}.

\paragraph{Multi.} The first model, $\text{Multi}$, is the classical multi-task learning framework with hard parameter sharing, where both tasks share the same BERT layers. Two dense layers for emotion classification and cause tagging operate at the same time from the same BERT layers, and we train both of the tasks simultaneously. That is, we simply calculate our emotion scores $e$ and cause tag scores $c$ from the same set of hidden states $H$.

We further develop two additional multi-task models with the intuition that we can design more explicit and concrete task dependencies than simple parameter sharing in the representation layer.

\paragraph{$\text{Multi}_{C \shortrightarrow E}$.} We assume that if a certain text span is given as the cause of an emotion, it should be possible to classify that emotion correctly while looking only at the words of the cause span. Therefore, we propose the $\text{Multi}_{C \shortrightarrow E}$ model, the architecture of which is illustrated in \autoref{fig:causefirst-arch}. This model begins with the single-task cause detection model, $BERT_C$, which produces a probability distribution  $P(y_i | x_i)$ over IOB tags for each token $x_i$, where $P(y_i | x_i) = c_i$ from \autoref{eqn:cause}. Then, for each token, we calculate the probability that it is part of the cause as $P(\text{Cause} | x_i) = P(B | x_i) + P(I | x_i) = 1 - P(O | x_i)$. We feed the resulting probabilities through a softmax over the sequence and use them as an attention distribution over the input tokens in order to pool the hidden representations and perform emotion classification: attention is computed as in \autoref{eqn:bahdanau-attn}, where $\alpha_i = \frac{\exp{P(\text{Cause} | x_i)}}{\sum_{j=1}^n \exp{P(\text{Cause} | x_i)}}$, and emotion classification as in \autoref{eqn:emotion}. For the $\text{Multi}_{C \shortrightarrow E}$ model, we apply teacher forcing at training time, and the gold cause spans are used to create the attention weights before emotion classification (which means that $P(\text{Cause} | x_i) \in \{0, 1\}$). At inference time, the model uses the predicted cause span instead. 

\paragraph{$\text{Multi}_{E \shortrightarrow C}$.} Next, we hypothesize that knowledge of the predicted emotion should help us identify salient cause words. The $\text{Multi}_{E \shortrightarrow C}$ model first performs emotion classification, which results in a probability distribution over predicted emotion labels, as in the $\text{BERT}_E$ model and \autoref{eqn:emotion}. We additionally keep an emotion embedding matrix $E$, where $E[i]$ is a learnable representation of the $i$-th emotion label (see \autoref{fig:emofirst-arch}) with dimension $d_e$ (in our experiments, we set $d_e = 300$). We use the predicted label probabilities $e$ to calculate a weighted sum of the emotion embeddings, i.e., $M = \sum_i e_i \cdot E[i]$. We then concatenate $M$ to the hidden representation of each token and perform emotion cause tagging with a final dense layer, i.e., $c_i = \text{softmax} (W_{c'}[h_i ; M] + b_{c'})$, where $;$ is the concatenation operator and $W_{c'} \in \mathbb{R}^{|\mathcal{C}| \times (d_{BERT} + d_e)}$ and $b_{c'} \in \mathbb{R}^{|\mathcal{C}|}$ are trainable parameters. In the $\text{Multi}_{E \shortrightarrow C}$ model, we again do teacher forcing and use the gold emotion labels before doing the sequence tagging for cause detection (i.e., $e$ is a one-hot vector where the gold emotion label has probability 1 and all other emotion labels have probability 0). At inference time, the model will use the predicted emotion distribution instead.

\subsection{Adapted Knowledge Models}

\begin{figure}
	\centering
	\includegraphics[scale=0.35]{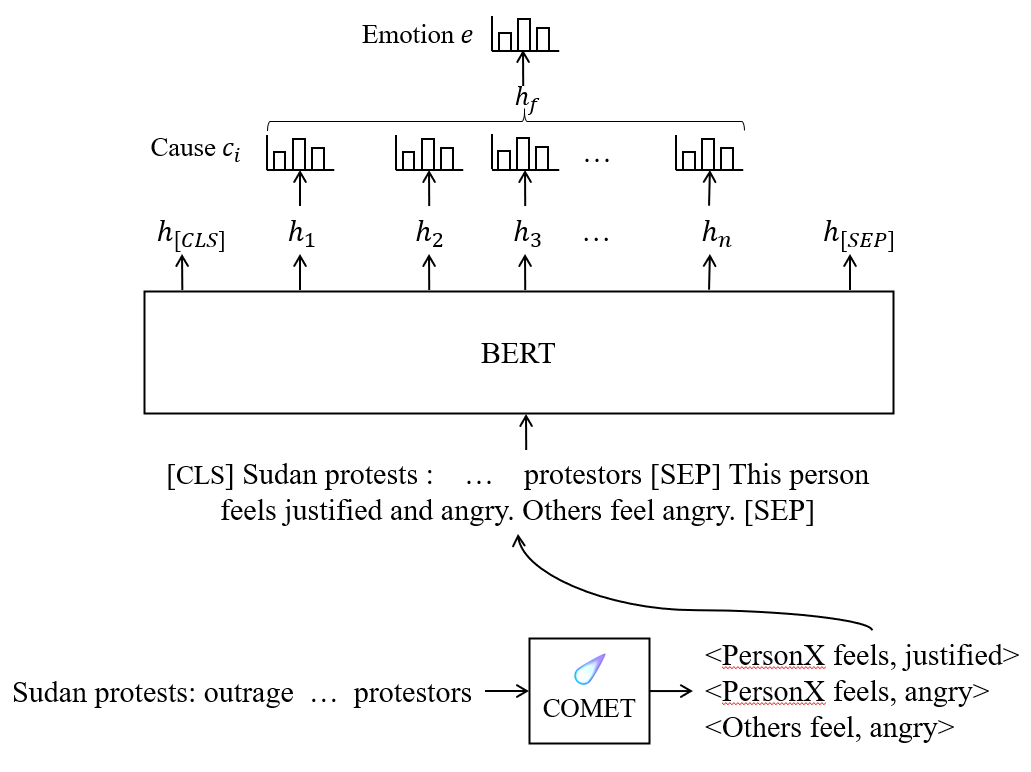}
	\caption{The architecture of our proposed $\text{Multi}^{COMET}_{C \shortrightarrow E}$ model.}
	\label{fig:comet-causefirst-architecture}
\end{figure}

\begin{table*}[t!]
	\centering
	\begin{tabular}{l|r|r|r}
		& \textbf{Emotion Macro F1} & \textbf{Emotion Accuracy} & \textbf{Cause Span F1} \\ \hline
		BERT & 37.25 $\pm$ 1.30 & 38.50 $\pm$ 0.84 & 37.49 $\pm$ 1.94 \\ \hline 
		$\text{BERT}^{COMET}$ & 37.74 $\pm$ 0.84 & 38.50 $\pm$ 1.14 & 39.27 $\pm$ 1.85 \\ \hline
		Multi & 36.91 $\pm$ 1.48 & 38.34 $\pm$ 1.94 & 38.35 $\pm$ 3.89 \\ \hline
		$\text{Multi}_{C \shortrightarrow E}$ & 37.74 $\pm$ 2.12 & 38.74 $\pm$ 2.07 & 39.08 $\pm$ 3.73 \\ \hline
		$\text{Multi}_{E \shortrightarrow C}$ & 38.26 $\pm$ 3.28 & 39.69 $\pm$ 3.41 & 38.83 $\pm$ 1.60 \\ \hline
		$\text{Multi}^{COMET}$ & 37.06 $\pm$ 2.04 & 39.05 $\pm$ 0.98 & \textbf{39.50} $\pm$ 2.25 \\ \hline
		$\text{Multi}_{C \shortrightarrow E}^{COMET}$ & \textbf{39.26}* $\pm$ 1.13 & \textbf{40.79} $\pm$ 2.17 & 38.68 $\pm$ 1.36 \\ \hline
		$\text{Multi}_{E \shortrightarrow C}^{COMET}$ & 37.44 $\pm$ 1.37 & 38.58 $\pm$ 1.44 & 36.27 $\pm$ 1.31 \\
	\end{tabular}
	\caption{The results of our models, averaged over five runs with the same five distinct random seeds. The model with the highest mean performance under each metric is bolded. Results marked with a * are statistically significant above the single-task BERT baseline by the paired t-test (p $<$ 0.05).}
	\label{tab:results}
\end{table*}

Recent work has shown that fine-tuning pre-trained language models such as GPT-2 on \emph{knowledge graph tuples} such as ConceptNet~\cite{li-etal-2016-commonsense} or ATOMIC \citep{DBLP:journals/corr/abs-1811-00146} allows these models to express their implicit knowledge directly \citep{DBLP:conf/acl/BosselutRSMCC19}. These adapted \emph{knowledge models} (e.g., COMET \citep{DBLP:conf/acl/BosselutRSMCC19}) can produce common-sense knowledge on-demand for any entity, relation or event.
Considering that common-sense knowledge plays an important role in understanding implicitly expressed emotions and the reasons for those emotions, we explore the use of common-sense knowledge for our tasks, in particular the use of COMET adaptively pre-trained on the ATOMIC event-centric knowledge base. ATOMIC's event relations include ``xReact'' and ``oReact'', which describe the feelings of certain entities after the input event occurs. For example, ATOMIC's authors present the example of $<$PersonX pays PersonY a compliment, xReact, PersonX will feel good$>$. xReact refers to the feelings of the primary entity in the event, and oReact refers to the feelings of others (in this instance, oReact yields ``PersonY will feel flattered''). For example, using the headline ``Sudan protests: Outrage as troops open fire on protestors", COMET-ATOMIC outputs that PersonX feels justified, PersonX feels angry, Others feel angry, and so on (\autoref{fig:comet-causefirst-architecture}). To use this knowledge model in our task, we modify our approach by reframing our single-sequence classification task as a sequence-pair classification task (for which BERT can be used directly). We feed our input headlines into COMET-ATOMIC (using the model weights released by the authors), collect the top two outputs for xReact and oReact using beam search decoding, and then feed them into BERT alongside the input headlines, as a second sequence using the SEP token. That is, our input to BERT is now $X = [[\text{CLS}], x_1, x_2, ..., x_n, [\text{SEP}], z_1, z_2, ..., z_m, [\text{SEP}]]$, where $z_i$ are the $m$ WordPiece tokens of our COMET output and are preprocessed in the same way as $x_i$. We hypothesize that, since pre-trained BERT is trained with a next sentence prediction objective, expressing the COMET outputs as a grammatical sentence will help BERT make better use of them, so we formulate this second sequence as complete sentences (e.g., ``This person feels... Others feel...'') (\autoref{fig:comet-causefirst-architecture}). 
 
This approach allows us incorporate information from  COMET into all our single- and multi-task BERT-based models; the example shown in  \autoref{fig:comet-causefirst-architecture} is our  $\text{Multi}_{C \shortrightarrow E}$ model. We refer to the COMET variants of these models as: $\text{BERT}^{COMET}$ (single-task models) and $\text{Multi}^{COMET}$, $\text{Multi}^{COMET}_{C \shortrightarrow E}$, $\text{Multi}^{COMET}_{E \shortrightarrow C}$ for the three multi-task models.

\section{Experimental Setup}

\paragraph{Evaluation Metrics}
For emotion classification, we report macro-averaged F1 and accuracy. For cause tagging, we report exact span-level F1 (which we refer to as \textit{span F1}), as developed for named entity recognition (e.g., \citet{tjong-kim-sang-de-meulder-2003-introduction}), where a span is marked as correct if and only if its type and span boundaries match the gold exactly\footnote{Our cause tagging task has only one type, ``cause'', as GoodNewsEveryone is aggregated such that each data point has exactly one emotion-cause pair. We note that this problem formulation leaves open the possibility of multiple emotion-cause pairs.}.

\paragraph{Training and Hyperparameter Selection} The classification layers are initialized randomly from a uniform distribution over $[-0.07,0.07]$\footnote{The default initialization from the \texttt{gluon} package: \url{https://mxnet.apache.org/versions/1.7.0/api/python/docs/api/gluon/index.html}}, and all the parameters are trained on our dataset for up to 20 epochs, with early stopping based on the performance on the validation data (macro F1 for emotion, span F1 for cause). All models are trained with the Adam optimizer \citep{DBLP:journals/corr/KingmaB14}. We highlight again that for our $\text{Multi}_{C \shortrightarrow E}$ and $\text{Multi}_{E \shortrightarrow C}$  models, we use teacher forced during training to avoid cascading training error. Because the subset of the data we use is relatively small, we follow current best practices for dealing with neural models on small data and select hyperparameters and models using the average performance of five models with different fixed random seeds on the development set. We then base our models' performance on the average of the results from these five runs (e.g., reported emotion F1 is the average of the emotion F1 scores for each of our five runs). For our joint models, since our novel models revolve around using one task as input for the other, we separately tune two sets of hyperparameters for each model, one based on each of the single-task metrics, yielding, for example, one Multi model optimized for predicting emotion and one optimized for predicting cause. The hyperparameters we tune are dropout in our linear layers, initial learning rate of the optimizer, COMET relation type, lambda weight for our multi-task models, and the type of pooler for emotion classification (enumerated in \autoref{sec:singletask}).

\section{Results} \label{sec:results}

We present the results of our models in \autoref{tab:results}\footnote{\citet{oberlander-klinger-2020-token} report an F1 score of 34 in this problem setting on this dataset, but on a larger subset of the data (as they do not limit themselves to the Ekman emotions) and so we cannot directly compare our work to theirs.}. We see that the overall best model for each task is a multi-task adapted knowledge model, with $\text{Multi}_{C \shortrightarrow E}^{COMET}$ performing best for emotion (which is a statistically significant improvement over BERT by the paired t-test, $p<0.05$) and $\text{Multi}^{COMET}$ performing best for cause. These results seem to support our two hypotheses: 1) emotion recognition and emotion cause detection can inform each other and 2) common-sense knowledge is helpful to infer the emotion and the cause for that emotion expressed in text.
Specifically, we notice that $\text{Multi}_{C \shortrightarrow E}$ alone does not outperform BERT on either cause or emotion, but $\text{Multi}_{C \shortrightarrow E}^{COMET}$ outperforms both BERT and $\text{Multi}_{C \shortrightarrow E}$ on both tasks. For cause, we also see additional benefits of common-sense reasoning alone: $\text{BERT}^{COMET}$ outperforms BERT (multi-task modeling alone, Multi, also outperforms BERT for this task) and $\text{Multi}^{COMET}$ outperforms Multi. These results speak to the differences between the two tasks, suggesting that common-sense reasoning, which aims to generates implicit emotions, and cause information may be complementary for emotion detection, but that for cause tagging, common-sense reasoning and given emotion information may overlap. 
The common-sense reasoning we have used in this task (xReact and oReact from ATOMIC) is expressed as possible emotional reactions to an input situation, so this makes intuitive sense.

\begin{figure}[t]
    \includegraphics[scale=0.55]{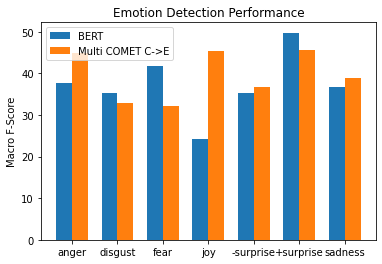}
    \caption{Performance of the BERT and $\text{Multi}_{C \shortrightarrow E}^{COMET}$ models on emotion classification.}
    \label{fig:emotion_breakdown}
\end{figure}

\begin{figure}[t]
	\includegraphics[scale=0.55]{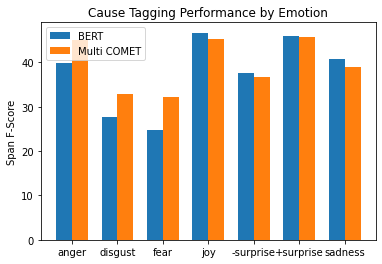}
	\caption{Performance of the BERT and $\text{Multi}^{COMET}$ models on cause tagging, broken down by emotion.}
	\label{fig:cause_breakdown}
\end{figure}

Finally, we also present per-emotion results for our best model for each task ($\text{Multi}_{C \shortrightarrow E}^{COMET}$ for emotion and $\text{Multi}^{COMET}$ for cause) against the single-task BERT baselines in \autoref{fig:emotion_breakdown} and \autoref{fig:cause_breakdown}; these per-emotion scores are again the average performance of models trained with each of our five random seeds. We see that each task improves on a different set of emotions: for emotion classification $\text{Multi}_{C \shortrightarrow E}^{COMET}$ consistently improves over BERT by a significant margin on joy and to a lesser extent on anger and sadness. Meanwhile, for cause tagging, $\text{Multi}^{COMET}$ improves over BERT on anger, disgust, and fear, while yielding very similar performance on the rest of the emotions. 

\section{Analysis and Discussion} \label{sec:analysis}

In order to understand the impact of common-sense reasoning and multi-task modeling for the two tasks, we provide several types of analysis in addition to our results in \autoref{sec:results}. First, we include examples of our various models' outputs showcasing the impact of our methods (\autoref{sec:examples}). Second, we carry out an analysis  of the dataset, focusing on the impact of label variation among multiple annotators  on the models' performance (\autoref{sec:label-analysis}).

\subsection{Example Outputs} \label{sec:examples}
\begin{table*}[t]
	\centering
	\begin{tabular}{|c|c|c|c|}
		\hline
		\textbf{BERT}                                                                      & \textbf{Multitask}                         & \textbf{$\text{BERT}^{COMET}$}                            & \textbf{$\text{Multitask}^{COMET}$}                         \\ \hline
		\multicolumn{4}{|c|}{\begin{tabular}[c]{@{}c@{}}Mexico reels from \hlcolor{shooting attack in El Paso}\\  \textbf{\textit{fear}}\end{tabular}}                                                                                                                \\ \hline
		negative surprise                                                                  & negative surprise                          & fear                                      & fear                                             \\ \hline
		\multicolumn{4}{|c|}{\begin{tabular}[c]{@{}c@{}}Insane video shows Viking Sky cruise \hlcolor{ship thrown into chaos at sea}\\ \textbf{\textit{fear}}\end{tabular}}                                                                                          \\ \hline
		negative surprise                                                                  & fear                                       & negative surprise                         & fear                                             \\ \hline
		\multicolumn{4}{|c|}{\begin{tabular}[c]{@{}c@{}}Durant \hlcolor{could return for Game 3}\\ \textbf{\textit{positive surprise}}\end{tabular}}                                                                                                                           \\ \hline
		for game                                                                           & \multicolumn{3}{|c|}{could return for game}                                                                                                 \\ \hline
		\multicolumn{4}{|c|}{\begin{tabular}[c]{@{}c@{}}Dan Fagan: \hlcolor{Triple shooting near New Orleans School yet another sign of city's crime problem}\\ \textbf{\textit{negative surprise}}\end{tabular}}                                   \\ \hline
		\begin{tabular}[c]{@{}c@{}}school yet another sign \\ of city's crime\end{tabular} & \multicolumn{3}{|c|}{\begin{tabular}[c]{@{}c@{}}: triple shooting near new orleans school yet another sign of city's \\ crime\end{tabular}} \\ \hline
	\end{tabular}
	\caption{Example outputs from our systems. For each example, the gold cause is highlighted in yellow and the gold emotion is given under the text; the first two examples give our models' emotion outputs; the latter two, their causes. Joined cells show that multiple models produced the same output. To make this table easier to read, ``Multitask'' here may refer to Multi, $\text{Multi}_{E \shortrightarrow C}$, or $\text{Multi}_{C \shortrightarrow E}$ (details on selection and results for each individual model available in appendix; most multi-task models gave similar outputs).}
	\label{tab:examples}
\end{table*}

\begin{table*}[t]
	\begin{adjustbox}{max width=\textwidth}
		\begin{tabular}{c|l|l|l|l|l|l|l|l}
			\textbf{Metric} & \textbf{BERT} & \textbf{$\text{BERT}^{COMET}$} & \textbf{Multi} & \textbf{$\text{Multi}_{E \shortrightarrow C}$} & \textbf{$\text{Multi}_{C \shortrightarrow E}$} & \textbf{$\text{Multi}^{COM}$} & \textbf{$\text{Multi}_{E \shortrightarrow C}^{COMET}$} & \textbf{$\text{Multi}_{C \shortrightarrow E}^{COMET}$} \\ \hline
			\begin{tabular}[c]{@{}l@{}}\textbf{Acc.}\\\textbf{(Gold)}\end{tabular} & 38.50 & 38.50 & 38.34 & 39.68 & 38.74 & 39.05 & 38.58 & 40.79 \\ \hline
			\begin{tabular}[c]{@{}l@{}}\textbf{Acc.}\\\textbf{($\neg$Gold)}\end{tabular}  & 23.48 & 23.24 & 22.37 & 21.11 & 22.85 & 21.26 & 22.45 & 20.08
		\end{tabular}
	\end{adjustbox}
	\caption{Comparison of gold accuracy and non-gold ($\neg$gold) accuracy for our emotion classification models.}
	\label{tab:annotator-comparison}
\end{table*}

We provide some example outputs from our systems for both cause and emotion in \autoref{tab:examples}; the various Multi models have been grouped together for readability and because they often produce similar outputs, but the outputs for every model are available in the appendix. In the first example, the addition of COMET to BERT informs the model enough to choose the gold emotion label; in the third and fourth, either COMET or multi-task learning is enough to help the model select key words that should be included in the cause (\textit{return} and \textit{triple shooting}). We also particularly note the second example, in which multi-task learning is needed both for the BERT and $\text{BERT}^{COMET}$ models to be able to correctly predict the gold emotion. This suggests that for cause, both common-sense reasoning and emotion classification may carry overlapping useful information for cause tagging, while for emotion, different instances may be helped more by different aspects of our models.

\subsection{Label Agreement} \label{sec:label-analysis}

Upon inspection of the GoodNewsEveryone data, we discover significant variation in the emotion labels produced by annotators as cautioned by the authors in their original publication\footnote{While the authors selected data according to agreement on the emotion labeling task, they found that in only 75\% of cases do at least 3 annotators agree, with diminishing rates of agreement for more annotators.}. 
From our inspection of the development data, we see recurring cases where different annotators give directly opposing labels for the same input, depending on how they interpret the headline and whose emotions they choose to focus on. For example, our development set includes the following example: \textit{Simona Stuns Serena at Wimbledon: Game, Set and ``Best Match'' for Halep}. The gold adjudicated emotion label for this example is \textit{negative surprise}, but annotators actually included multiple primary and secondary emotion labels including \textit{joy}, \textit{negative surprise}, \textit{positive surprise}, \textit{pride}, and \textit{shame}, which can be understood as various emotions felt by the two entities participant in the event (Simona Halep and Serena Williams). For this input, COMET suggests xReact may be \textit{happy} or \textit{proud} and oReact may be \textit{happy} --- these reactions are likely most appropriate for tennis player Simona Halep, but not the only possible emotion that can be inferred from the headline.

Inspired by the variation in the data, we compute also models' accuracy using the human annotations that did not agree with the gold (i.e., a predicted emotion label is correct if it was  suggested by a human annotator but was not part of a majority vote to be included in the gold). We denote this $\neg$Gold, and we compare the performance of our models with respect to Gold and $\neg$Gold. We present the results of this analysis in \autoref{tab:annotator-comparison}\footnote{Note that we perform this analysis on just one of our five runs of the model, so the accuracy numbers do not exactly correspond to those in \autoref{tab:results}.}. 
In this table, a higher $\neg$Gold accuracy means that the model is more likely to produce emotion labels that were not the gold but were suggested by some annotator.  First of all, we note that all models have a relatively high $\neg$Gold accuracy (about half the magnitude of their gold accuracy); we believe this reflects the wide variety of annotations given by the annotators. We see a tradeoff between the Gold and $\neg$Gold accuracy, and we note that generally the single-task models have higher $\neg$Gold accuracy and the COMET-enhanced multi-task models have higher Gold accuracy. This suggests that our language models have general knowledge about emotion already, but that applying common-sense knowledge helps pare down the space of plausible outputs to those that are most commonly selected by human annotators. Recall that this dataset was annotated by taking the most frequent of the annotator-provided emotion labels.
Further, since the multi-task models have higher Gold accuracy and lower $\neg$Gold accuracy than the single-task models, this suggests that also predicting the cause of an emotion causes the model to narrow down the space of possible emotion labels to only those that are most common.

\section{Conclusions and Future Work}

We present a common-sense knowledge-enhanced multi-task framework for joint emotion detection and emotion cause tagging. Our inclusion of common-sense reasoning through COMET, combined with multi-task learning, yields performance gains on both tasks including significant gains on emotion classification.
We highlight the fact that this work frames the cause extraction task as a span tagging task, allowing for the future possibility of including multiple emotion-cause pairs per input or multiple causes per emotion and allowing the cause to take on any grammatical role. Finally, we present an analysis of our dataset and models, showing that labeling emotion and its semantic roles is a hard task with annotator variability, 
but that common-sense knowledge helps language models focus on the most prominent emotions according to human annotators. In future work, we hope to explore ways to integrate common-sense knowledge more innately into our classifiers and ways to apply these models to other fine-grained emotion tasks such as detecting the experiencer or the target of an emotion.

\section*{Acknowledgements}

We would like to thank our reviewers as well as the members of Amazon AI team for their constructive and insightful feedback.

\section*{Ethical Considerations}

Our intended use for this work is as a tool to help understand emotions expressed in text. We propose that it may be useful for things like product reviews (where producers and consumers can rapidly assess reviews for aspects of their products to improve or expand), disaster relief (where those in need of help from any type of disaster can benefit if relief agents can understand what events are causing negative emotions, during and after the initial disaster), and policymaking (where constituents can benefit if policymakers can see real data about what policies are helpful or not and act in their interests). These applications do depend on the intentions of the user, and a malicious actor may certainly misuse the ability to (accurately or inaccurately) detect emotions and their causes. We do not feel it responsible to publicly list the ways in which this may happen in this paper. We also believe that regulators and operators of this technology should be aware that it is still in its nascent stages and does not represent an infallible oracle --- the predictions of this and any model should be reviewed by humans in the loop, and we feel that general public awareness of the limitations and mistakes of these models may help mitigate any possible harm. If these models are inaccurate, they will output either the incorrect emotion or the incorrect cause; blindly trusting the model's predictions without examining them may lead to unfair consequences in any of the above applications (e.g., failure to help someone whose text is misclassified as positive surprise during a natural disaster or a worsened product or policy if causes are incorrectly predicted). We additionally note that in its current form, this work is intended to detect the emotions that are expressed in text (headlines), and not those of the reader.

We concede that the data used in this work consists of news headlines and may not be the most adaptable to the use cases we describe above; we caution that models trained on these data will likely require domain adaptation to perform well in other settings. \citet{DBLP:conf/lrec/BostanKK20} report that their data comes from the Media Bias Chart\footnote{\url{https://www.adfontesmedia.com/about-the-interactive-media-bias-chart/}}, which reports that their news sources contain a mix of political views, rated by annotators who also self-reported a mix of political views. We note that these data are all United States-based and in English. \citet{DBLP:conf/lrec/BostanKK20} do sub-select the news articles according to impact on Twitter and Reddit, which have their own user-base biases\footnote{\url{https://www.pewresearch.org/internet/fact-sheet/social-media/}}, typically towards young, white American men; therefore, the data is more likely to be relevant to these demographics. The language used in headlines will likely most resemble Standard American English as well, and therefore our models will be difficult to use directly on other dialects and vernaculars.

\bibliography{custom}
\bibliographystyle{acl_natbib}

\clearpage
\appendix

\section{Appendix}

\subsection{Hyperparameter Tuning}

\begin{table*}[ht]
	\centering
	\begin{tabular}{l|l|l}
		Parameter Name & Type & Range or Values \\ \hline
		\texttt{pooler} & Categorical & \texttt{[cls, mean, max, attention]} \\
		\texttt{learning rate} & Continuous & $[10^{-6}, 10^{-4}]$ \\
		\texttt{dropout} & Continuous & $[0, 0.9]$ \\
		\texttt{lambda} & Continuous & $[0.1, 0.9]$ \\
		\texttt{comet\_relations} & Categorical & \texttt{[xReact, oReact, both]}\\
	\end{tabular}
	\caption{Our hyperparameter search ranges.}
	\label{tab:hyperparams}
\end{table*}

\begin{table*}[ht]
	\centering
	\begin{tabular}{c|c|l|l}
		\multicolumn{1}{l|}{\textbf{Model}} & \textbf{Target Task} & \textbf{Parameter Name} & \textbf{Parameter Value} \\ \hline
		\multirow{3}{*}{$\text{BERT}_E$} & \multirow{3}{*}{Emotion} & pooler & \texttt{cls} \\
		& & dropout & 0.8999992513311351 \\
		& & lr & $2.0872134970009262 \times 10^{-5}$ \\ \hline
		\multirow{2}{*}{$\text{BERT}_C$} & \multirow{2}{*}{Cause} & dropout & 0.04011659404129298 \\
		& & lr & $9.609926650689472 \times 10^{-5}$ \\ \hline
		\multirow{4}{*}{$\text{BERT}_E^{COMET}$} & \multirow{4}{*}{Emotion} & pooler & \texttt{cls} \\
		& & dropout & 0.6467089448672897 \\
		& & lr & $3.548213539029209 \times 10^{-5}$ \\
		& & comet\_relations & \texttt{both} \\ \hline
		\multicolumn{1}{l|}{\multirow{3}{*}{$\text{BERT}_C^{COMET}$}} & \multirow{3}{*}{Cause} & dropout & 0.8806119007595122 \\
		\multicolumn{1}{l|}{} & & lr & $9.913585728926367 \times 10^{-5}$ \\
		\multicolumn{1}{l|}{} & & comet\_relations & \texttt{xReact}                        
	\end{tabular}
	\caption{The selected hyperparameter values for our single-task models.}
	\label{tab:hyperparam-values-singletask}
\end{table*}

\begin{table*}[ht]
	\centering
	\begin{tabular}{c|c|l|l}
		\textbf{Model}                            & \textbf{Target Task}     & \textbf{Parameter Name} & \textbf{Parameter Value} \\ \hline
		\multirow{8}{*}{Multi}                    & \multirow{4}{*}{Emotion} & pooler                  & \texttt{mean}                         \\
		&                          & dropout                 & 0.1438975482079587                         \\
		&                          & lr                      & $2.170218150294524 \times 10^{-5}$                         \\ 
		&                          & lambda                  & 0.3736515054477897                          \\ \cline{2-4} 
		& \multirow{4}{*}{Cause}   & pooler                  & \texttt{cls}                         \\
		&                          & dropout                 & 0.8929935089177194                         \\
		&                          & lr                      & $9.929740332732521 \times 10^{-5}$                         \\ 
		&                          & lambda                  & 0.6103686494768474                         \\ \hline
		
		\multirow{8}{*}{$\text{Multi}_{E \shortrightarrow C}$}                    & \multirow{4}{*}{Emotion} & pooler                    & \texttt{max}                         \\
		&                          & dropout                 & 0.2511612834815036                         \\
		&                          & lr                      & $3.179072019077849 \times 10^{-5}$                         \\
		&                          & lambda                  & 0.4938386162506444                         \\ \cline{2-4} 
		& \multirow{4}{*}{Cause}   & pooler                  & \texttt{max}                         \\
		&                          & dropout                 & 0.763419047616446                         \\
		&                          & lr                      & $8.680439371509037 \times 10^{-5}$                         \\
		&                          & lambda                  & 0.1341940851689314                         \\ \hline		\multirow{8}{*}{$\text{Multi}_{C \shortrightarrow E}$} & \multirow{4}{*}{Emotion} & pooler                  & \texttt{max}                         \\
		&                          & dropout                 & 0.8138762283528274                         \\
		&                          & lr                      & $4.2586257586160994 \times 10^{-5}$                         \\
		&                          & lambda                  & 0.8531247637209994                         \\ \cline{2-4}
		& \multirow{4}{*}{Cause}   & pooler                  & \texttt{mean}                         \\
		&                          & dropout                 & 0.6992099059226856                         \\
		&                          & lr                      & $9.859155309987275 \times 10^{-5}$                         \\
		&                          & lambda                  & 0.4855821360212248                
	\end{tabular}
	\caption{The selected hyperparameter values for our multi-task BERT models.}
	\label{tab:hyperparam-values-bert-multitask}
\end{table*}

\begin{table*}[ht]
	\centering
	\begin{tabular}{c|c|l|l}
		\textbf{Model}                            & \textbf{Target Task}     & \textbf{Parameter Name} & \textbf{Parameter Value} \\ \hline
		\multirow{10}{*}{$\text{Multi}^{COMET}$}                    & \multirow{5}{*}{Emotion} & pooler                  & \texttt{max}                         \\
		&                          & dropout                 & 0.22350077887111716                          \\
		&                          & lr                      & $3.137385699389837 \times 10^{-5}$                         \\ 
		&                          & lambda                  & 0.7676911585403968                         \\ 
		&                          & comet\_relations        & \texttt{both}                         \\ \cline{2-4} 
		& \multirow{5}{*}{Cause}   & pooler                  & \texttt{mean}                         \\
		&                          & dropout                 & 0.8891347000216091                         \\
		&                          & lr                      & $8.123006047625093 \times 10^{-5}$                         \\ 
		&                          & lambda                  & 0.1                         \\ 
		&                          & comet\_relations        & \texttt{both}                         \\ \hline	
		\multirow{10}{*}{$\text{Multi}_{E \shortrightarrow C}^{COMET}$}                    & \multirow{5}{*}{Emotion} & pooler                    & \texttt{mean}                         \\
		&                          & dropout                 & 0.1372637910712323                         \\
		&                          & lr                      & $3.0408118480380588 \times 10^{-5}$                         \\
		&                          & lambda                  & 0.8968243966922735                         \\ 
		&                          & comet\_relations        & \texttt{both}                         \\ \cline{2-4} 
		& \multirow{5}{*}{Cause}   & pooler                  & \texttt{max}                         \\
		&                          & dropout                 & 0.5319636087561394                         \\
		&                          & lr                      & $7.581334242472624 \times 10^{-5}$                         \\
		&                          & lambda                  & 0.10896064677810494                         \\ 
		&                          & comet\_relations        & \texttt{both}                         \\ \hline		
		\multirow{10}{*}{$\text{Multi}_{C \shortrightarrow E}^{COMET}$} & \multirow{5}{*}{Emotion} & pooler                  & \texttt{cls}                         \\
		&                          & dropout                 & 0.7359624181177503                         \\
		&                          & lr                      & $1.9853909769532754 \times 10^{-5}$                         \\
		&                          & lambda                  & 0.7947522633173147                         \\ 
		&                          & comet\_relations        & \texttt{both}                         \\ \cline{2-4}
		& \multirow{5}{*}{Cause}   & pooler                  & \texttt{max}                         \\
		&                          & dropout                 & 0.01896406469706125                         \\
		&                          & lr                      & $8.360862387915605 \times 10^{-5}$                         \\
		&                          & lambda                  & 0.14588492191321054                         \\
		&                          & comet\_relations        & \texttt{oReact}                      
	\end{tabular}
	\caption{The selected hyperparameter values for our multi-task COMET models.}
	\label{tab:hyperparam-values-comet-multitask}
\end{table*}

We include descriptions of our hyperparameter tuning setup and the selected hyperparmeters for each of our models in \autoref{tab:hyperparams}; we note that single-task cause models ($\text{BERT}_C$ and $\text{COMET}_C$) do not tune the \texttt{pooler}, all single-task models do not tune the lambda parameter, and all non-common-sense models do not tune \texttt{comet\_relations}. The parameters selected by all of our models can be seen in \autoref{tab:hyperparam-values-singletask}, \autoref{tab:hyperparam-values-bert-multitask}, and \autoref{tab:hyperparam-values-comet-multitask}. All of our models are trained with minibatches of size $b = 32$.

We used Bayesian optimization as implemented by Amazon SageMaker\footnote{\url{https://aws.amazon.com/sagemaker/}} to tune these parameters, giving the learning rate a logarithmic scale and the dropout and lambda a linear one and allowing 75 iterations of parameter choice before selecting the setting with the best performance on the development set. Each individual instance of each model consisted of five different restarts with five distinct random seeds; one of these instances took approximately five minutes on a single Tesla V100 GPU, for a total of about 6.25 GPU-hours per model and thus 87.5 GPU-hours overall (since each multi-task model was trained twice: once optimized for emotion and once optimized for cause).

\subsection{Model Sizes}
Our models' sizes are dominated by BERT-base, which has 110 million trainable parameters \citep{DBLP:conf/naacl/DevlinCLT19}. We note that our trainable dense layers that interface with BERT have sizes 768 $\times$ 7 for emotion classification, 768 $\times$ 3 for cause tagging, and 1068 $\times$ 7 for our $\text{Multi}_{E \shortrightarrow C}$ models, while our emotion embedding matrix $E$ has 300 $\times$ 7 trainable parameters. Our fine-tuning process does continue to tune all of BERT's parameters.

\subsection{Extended Examples}

\begin{table*}
	\centering
	\begin{tabular}{lc}
		\multicolumn{2}{c}{\begin{tabular}[c]{@{}c@{}}Mexico reels from \hlcolor{shooting attack in El Paso}\\  \textbf{\textit{fear}}\end{tabular}} \\ \hline
		\multicolumn{1}{l|}{Model}                           & Output                        \\ \hline
		\multicolumn{1}{l|}{BERT}                            & negative surprise                              \\
		\multicolumn{1}{l|}{$\text{BERT}^{COMET}$}                           & fear                              \\
		\multicolumn{1}{l|}{Multi}                           & negative surprise                              \\
		\multicolumn{1}{l|}{$\text{Multi}_{C \shortrightarrow E}$}                         & negative surprise                               \\
		\multicolumn{1}{l|}{$\text{Multi}_{E \shortrightarrow C}$}                         & negative surprise                              \\
		\multicolumn{1}{l|}{$\text{Multi}^{COMET}$}                         & fear                              \\
		\multicolumn{1}{l|}{$\text{Multi}_{C \shortrightarrow E}^{COMET}$}                         & fear                              \\
		\multicolumn{1}{l|}{$\text{Multi}_{E \shortrightarrow C}^{COMET}$}                         & fear                             
	\end{tabular}
	\caption{Full model outputs for our first provided example.}
	\label{tab:full-example-1}
\end{table*}

\begin{table*}
	\centering
	\begin{tabular}{lc}
		\multicolumn{2}{c}{\begin{tabular}[c]{@{}c@{}}Insane video shows Viking Sky cruise \hlcolor{ship thrown into chaos at sea}\\ \textbf{\textit{fear}}\end{tabular}} \\ \hline
		\multicolumn{1}{l|}{Model}                           & Output                        \\ \hline
		\multicolumn{1}{l|}{BERT}                            & negative surprise                               \\
		\multicolumn{1}{l|}{$\text{BERT}^{COMET}$}                           & negative surprise                              \\
		\multicolumn{1}{l|}{Multi}                           & fear                              \\
		\multicolumn{1}{l|}{$\text{Multi}_{C \shortrightarrow E}$}                         & negative surprise                               \\
		\multicolumn{1}{l|}{$\text{Multi}_{E \shortrightarrow C}$}                         & fear                                \\
		\multicolumn{1}{l|}{$\text{Multi}^{COMET}$}                         & fear                               \\
		\multicolumn{1}{l|}{$\text{Multi}_{C \shortrightarrow E}^{COMET}$}                         & fear                              \\
		\multicolumn{1}{l|}{$\text{Multi}_{E \shortrightarrow C}^{COMET}$}                         & negative surprise                             
	\end{tabular}
	\caption{Full model outputs for our second provided example.}
	\label{tab:full-example-2}
\end{table*}

\begin{table*}
	\centering
	\begin{tabular}{lc}
		\multicolumn{2}{c}{\begin{tabular}[c]{@{}c@{}}Durant \hlcolor{could return for Game 3}\\ \textbf{\textit{positive surprise}}\end{tabular}} \\ \hline
		\multicolumn{1}{l|}{Model}                           & Output                        \\ \hline
		\multicolumn{1}{l|}{BERT}                            & for game                              \\
		\multicolumn{1}{l|}{$\text{BERT}^{COMET}$}                           & could return for game                              \\
		\multicolumn{1}{l|}{Multi}                           & could return for game                              \\
		\multicolumn{1}{l|}{$\text{Multi}_{C \shortrightarrow E}$}                         & could return for game                               \\
		\multicolumn{1}{l|}{$\text{Multi}_{E \shortrightarrow C}$}                         & could return for game                              \\
		\multicolumn{1}{l|}{$\text{Multi}^{COMET}$}                         & could return for game                               \\
		\multicolumn{1}{l|}{$\text{Multi}_{C \shortrightarrow E}^{COMET}$}                         & could return for game                              \\
		\multicolumn{1}{l|}{$\text{Multi}_{E \shortrightarrow C}^{COMET}$}                         & could return for game                              
	\end{tabular}
	\caption{Full model outputs for our third provided example.}
	\label{tab:full-example-3}
\end{table*}

\begin{table*}
	\centering
	\begin{tabular}{lc}
		\multicolumn{2}{c}{\begin{tabular}[c]{@{}c@{}}Dan Fagan: \hlcolor{Triple shooting near New Orleans School yet another sign of city's crime problem}\\ \textbf{\textit{negative surprise}}\end{tabular}} \\ \hline
		\multicolumn{1}{l|}{Model}                           & Output                        \\ \hline
		\multicolumn{1}{l|}{BERT}                            & school yet another sign of city's crime                               \\
		\multicolumn{1}{l|}{$\text{BERT}^{COMET}$}                           & : triple shooting near new orleans school yet another sign of city's crime                              \\
		\multicolumn{1}{l|}{Multi}                           & shooting near new orleans school yet another sign of city's crime                              \\
		\multicolumn{1}{l|}{$\text{Multi}_{C \shortrightarrow E}$}                         & : triple shooting near new orleans school yet another sign of city's crime                               \\
		\multicolumn{1}{l|}{$\text{Multi}_{E \shortrightarrow C}$}                         & : triple shooting near new orleans school yet another sign of city's crime                              \\
		\multicolumn{1}{l|}{$\text{Multi}^{COMET}$}                         & : triple shooting near new orleans school yet another sign of city's crime                              \\
		\multicolumn{1}{l|}{$\text{Multi}_{C \shortrightarrow E}^{COMET}$}                         & : triple shooting near new orleans school yet another sign of city's crime                              \\
		\multicolumn{1}{l|}{$\text{Multi}_{E \shortrightarrow C}^{COMET}$}                         & : triple shooting near new orleans school yet another sign of city's crime                             
	\end{tabular}
	\caption{Full model outputs for our fourth provided example.}
	\label{tab:full-example-4}
\end{table*}

We include the output of all models for our four selected examples in \autoref{sec:examples} in \autoref{tab:full-example-1}, \autoref{tab:full-example-2}, \autoref{tab:full-example-3}, and \autoref{tab:full-example-4}.

\end{document}